\documentclass[twoside,leqno,twocolumn]{article}

\usepackage[letterpaper]{geometry}

\usepackage{ltexpprt}
\usepackage{hyperref}

\usepackage{graphicx}
\usepackage{subcaption}

\begin{document}

\newcommand\relatedversion{}
\renewcommand\relatedversion{\thanks{The full version of the paper can be accessed at \protect\url{https://arxiv.org/abs/1902.09310}}} 

\title{Data-centric AI: Perspectives and Challenges}
\author{Daochen Zha\thanks{Rice University, \{daochen.zha,khlai,fyang,Xia.Hu\}@rice.edu}
\and Zaid Pervaiz Bhat\thanks{Texas A\&M University, zaid.bhat1234@tamu.edu}
\and Kwei-Herng Lai\footnotemark[1]
\and Fan Yang\footnotemark[1]
\and Xia Hu\footnotemark[1]}

\date{}

\maketitle


\fancyfoot[R]{\scriptsize{Copyright \textcopyright\ 2023 by SIAM\\
Unauthorized reproduction of this article is prohibited}}





\begin{abstract}

The role of data in building AI systems has recently been significantly magnified by the emerging concept of data-centric AI (DCAI), which advocates a fundamental shift from model advancements to ensuring data quality and reliability. Although our community has continuously invested efforts into enhancing data in different aspects, they are often isolated initiatives on specific tasks. To facilitate the collective initiative in our community and push forward DCAI, we draw a big picture and bring together three general missions: training data development, inference data development, and data maintenance. We provide a top-level discussion on representative DCAI tasks and share perspectives. Finally, we list open challenges. More resources are summarized at \url{https://github.com/daochenzha/data-centric-AI}
\end{abstract}


\section{Introduction}
\label{sec:1}

\begin{figure}[t]
  \centering
  \begin{subfigure}[b]{0.19\textwidth}
    \centering
    \includegraphics[width=0.99\textwidth]{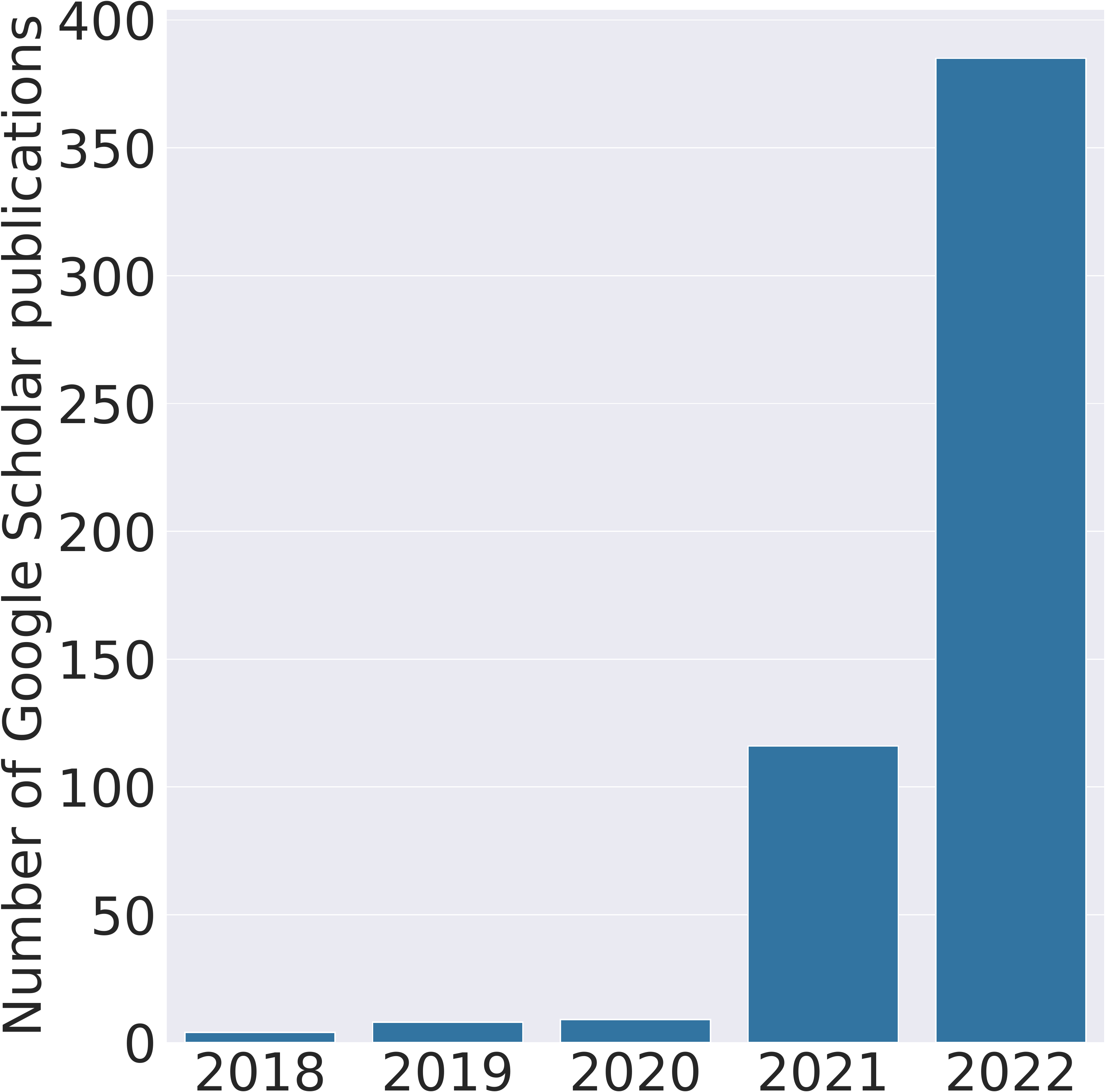}
  \end{subfigure}%
  \begin{subfigure}[b]{0.29\textwidth}
    \centering
    \includegraphics[width=0.99\textwidth]{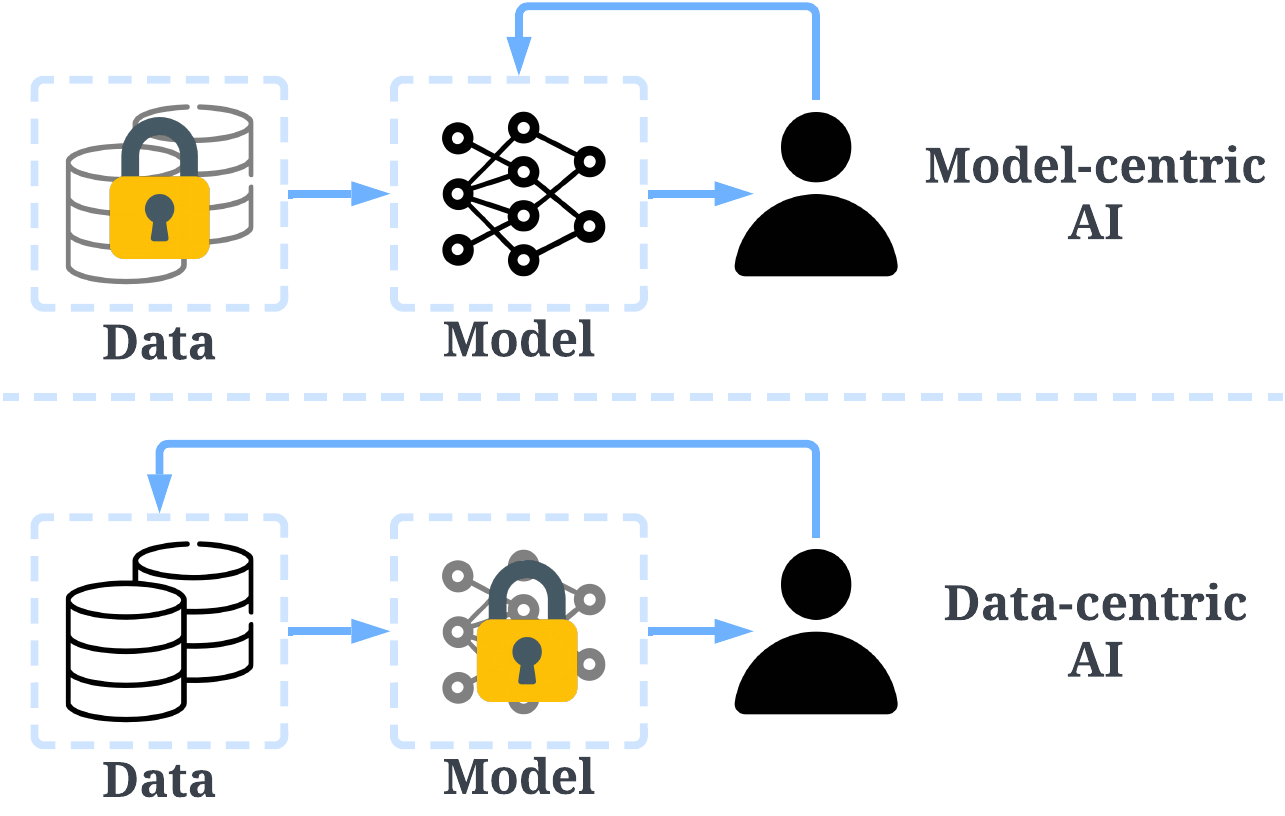}
  \end{subfigure}%
  \vspace{-5pt}
  \caption{\textbf{Left:} Tendency of DCAI over the past five years. The statistics are collected by querying Google Scholar with the exactly matched phrase ``data-centric AI". \textbf{Right:} Model-centric AI versus DCAI.}
  \label{fig:1}
  \vspace{-10pt}
\end{figure}

Data is an indispensable element of AI systems. Recently, its role has been significantly magnified by the emerging concept of data-centric AI (DCAI), shown on the left of Figure~\ref{fig:1}. The popularity of DCAI is mainly driven by a campaign launched by Ng et al.~\cite{ng2021data}, which advocates for a more data-centric rather than model-centric strategy for machine learning, a fundamental shift from model design to data quality and reliability.

The right-hand side of Figure~\ref{fig:1} illustrates a high-level comparison between model-centric AI and DCAI. In the conventional model-centric lifecycle, the benchmark dataset remains mostly unchanged. The primary goal of the researchers and practitioners is to iterate the model to improve performance. Although this paradigm encourages model advancements, it trusts too much in data; however, data could be susceptible to undesirable flaws, which raises a question: \emph{can the model performance reflect the actual capability, or is it just overfitting the dataset?} ``Garbage in, garbage out" is often one of the first lessons we have learned in machine learning. That being said, data is not merely a fuel for AI but rather a determining factor of the model quality. DCAI has become a recent trend to shift our focus from models toward data. Attention to data in our community can help build more powerful AI systems to deal with more complex real-world problems.

\begin{figure*}[t]
  \centering
    \includegraphics[width=1.0\textwidth]{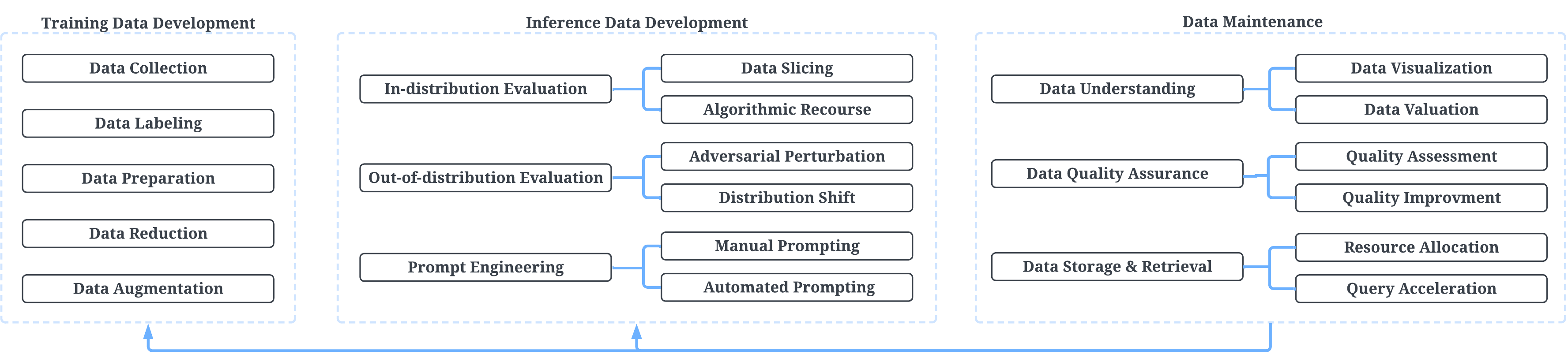}
  \vspace{-7.5pt}
  \caption{A big picture of DCAI missions and their representative tasks/subtasks.}
  \vspace{-10.5pt}
  \label{fig:overview}
\end{figure*}

In this paper, we define DCAI as a class of systematic techniques that develop, iterate, and maintain data for AI systems. While DCAI appears to be a new concept~\cite{polyzotis2021can,jarrahi2022principles,jakubik2022data}, many relevant research topics are not new. Our community has continuously invested efforts into enhancing data in different aspects, especially training data development. For instance, data augmentation~\cite{feng2021survey} has been extensively investigated to improve data diversity by adding slightly modified data samples or new synthetic samples into training sets; feature selection~\cite{li2017feature} has been studied since decades ago for preparing cleaner and more understandable data.

Despite these individual initiatives on specific tasks, there is a lack of a top-level summary and outlook of DCAI. In particular, the role of data stretches well beyond constructing data for training. First, it is comparably crucial to build a novel evaluation set to assess and understand the model quality comprehensively. Second, in industrial applications, data is not created once but rather demands continuous maintenance. It is essential to develop efficient algorithms, tools, and infrastructures to understand and debug data.

To facilitate the collective initiative in our community and push forward DCAI, we draw a big picture and share our perspectives with peers. Guided by Figure~\ref{fig:overview}, our discussion covers three general DCAI missions: training data development (Section~\ref{sec:2}), inference data development (Section~\ref{sec:3}), and data maintenance (Section~\ref{sec:4}). Then, we discuss open challenges in Section~\ref{sec:5}, followed by a conclusion in Section~\ref{sec:6}.

\section{Training Data Development}
\label{sec:2}



Training data is a collection of data instances used to teach machine learning models. Constructing high-quality training data is critical to achieving DCAI. In general, there are many tasks in the data construction process, which can be divided into five high-level goals: 1) data collection, 2) data labeling, 3) data preparation, 4) data reduction, and 5) data augmentation.




\textbf{Data collection} is the process of gathering data. A straightforward approach is to construct new datasets from scratch. However, this is often time-consuming. Thus, more efficient methods are proposed by leveraging the existing datasets. A representative task is \emph{dataset discovery}~\cite{bogatu2020dataset}, which aims to identify the most relevant datasets from a data lake (a repository of data stored in its raw format). \emph{Data integration} is another task that combines multiple datasets into a larger dataset~\cite{stonebraker2018data}. As more and more datasets become available, we anticipate the popularity of amassing datasets of interest will grow, and so does the need for more scalable algorithms.

\textbf{Data labeling} aims to add informative labels to data samples. Manual methods, such as \emph{crowdsourcing}, are accurate but costly. \emph{Semi-supervised labeling} is a family of techniques that infer the labels of unlabeled data based on a small amount of labeled data~\cite{xu2021dp,karamanolakis2021self} (e.g., training a model to make predictions on the unlabeled data). \emph{Active learning} is an iterative labeling strategy that selects the most informative unlabeled samples in each iteration~\cite{ren2021survey,zha2020meta,dong2023active}. Other research has redefined the labeling procedure in weakly-supervised settings~\cite{ratner2016data,zha2019multi}. For example, \emph{data programming}~\cite{ratner2016data}, takes domain-specific heuristics as input to infer labels. Large labeled datasets are key enablers of deep learning. We predict that more efficient labeling methods with various formats of human involvement for diverse data types will be proposed.

\textbf{Data preparation} cleans and transforms raw data into a form suitable for learning. \emph{Data cleaning} helps remove noises and errors in data~\cite{chu2016data}, such as imputing missing values, dropping duplicates, and fixing inconsistencies. \emph{Feature extraction} aims to create numerical features usable for learning from raw data~\cite{salau2019feature}. The extraction strategies often depend on data formats (e.g., tf–idf for text and patches for images). \emph{Data transformation} techniques, such as standardization and normalization~\cite{zha2022towards2}, make data more suitable for model training. Our anticipation for data preparation is that future research will focus more on automatically identifying the best strategies to deal with arbitrary data.

\textbf{Data reduction} reduces the data size, making it simpler and more understandable with potentially improved performance. From the feature perspective, common strategies include selecting a subset of features (\emph{feature selection}~\cite{li2017feature}), transforming data to a lower-dimension space (\emph{dimension reduction}~\cite{fodor2002survey}), etc. From the instance perspective, \emph{instance selection} is a notable research topic that aims to select a subset of the data that maintains the underlying distribution~\cite{olvera2010review}. As modern datasets are becoming increasingly larger (both feature and instance sizes), data reduction plays a critical role in performance and capacity optimization.


\textbf{Data augmentation} is a strategy to increase data diversity by creating modified samples without collecting more data. The existing methods can generally be divided into \emph{basic manipulation} and \emph{deep learning} approaches~\cite{shorten2019survey,zha2022towards}. The former directly makes minor changes to the original data samples, e.g., flipping images. The latter trains deep learning models to synthesize data, e.g., training generative models to capture data distributions and generate samples. Data augmentation brings many benefits, such as improved accuracy, increased generalization capabilities, and robustness.

We recognize that more research efforts have been paid to data processing (data reduction and augmentation) than data creation (data collection and labeling). This could be partly explained by our focus on model-centric AI in the past since processing data serves as a pre-processing step of model training. Looking forward, we anticipate that more studies on data creation will appear since it fundamentally determines data quality.


\section{Inference Data Development}
\label{sec:3}



In the model-centric paradigm, inference data typically refers to test and validation sets where some metrics (e.g., accuracy) are obtained to measure prediction performances. In the DCAI paradigm, we target a more comprehensive model evaluation with more granular insights into model capabilities beyond performance metrics, which includes \textit{in-distribution} and \textit{out-of-distribution} evaluation data. Additionally, with the advent of large language models, \emph{prompt engineering} has emerged as a promising approach to probe knowledge from a model without updating the model itself.


\textbf{In-distribution evaluation} data refers to testing samples that follow the same distribution as the training data. Some techniques have been proposed to create highly specific in-distribution sets for various testing purposes. One representative technique is \textit{data slicing}~\cite{chung2019slice,sagadeeva2021sliceline}, which reserves a small part of original data based on populations (e.g., genders or races). This can provide important information about where models underperform (e.g., a model could be less accurate for females than males). Another technique \textit{Algorithmic recourse}~\cite{venkatasubramanian2020philosophical,karimi2021algorithmic}, which is an emerging task to understand the decision boundaries of models. It aims to find a hypothetical in-distribution set close to the model boundary but with different predictions. For instance, if an individual is denied a loan, algorithmic recourse seeks a close sample (e.g., with an increased account balance) that can be accepted. It helps explain predictions and prevent ethical issues and fatal errors.

\textbf{Out-of-distribution evaluation} data means the testing samples follow a distribution that differs from the training data. Numerous techniques are employed for various evaluation purposes. To test robustness, a typical approach is \textit{adversarial perturbation}, which aims to design samples similar to the original ones but can mislead the model~\cite{goodfellow2014explaining}. The adversarial sets can reveal the high-risk weakness, provide informative features to refine the model training and evaluate ethical issues. Furthermore, To assess model transferability, we can purposely develop data to evaluate the adaptation on \emph{distribution shift}~\cite{sugiyama2007covariate}, a common issue for model deployment. This is often achieved by simulating the distribution shift through weighted sampling.

\textbf{Prompt engineering} is a rapidly developing field that involves designing prompts to unlock model capabilities, which involves: manual prompting, which constructs prompts based on pre-defined templates~\cite{schick2020few}, and automated prompting, which searches or learns the best prompts to achieve a given goal~\cite{gao2021making}.

Unlike training data construction, research on inference data development is relatively open-ended. The strategies are often purpose-driven, aiming to test a specific property or unlock a certain capability of the model. We predict that more systematic research will be conducted in this direction, and more inference data creation algorithms will be proposed.

\section{Data Maintenance}
\label{sec:4}



In production scenarios, data is not created once, but rather continuously updated. Data maintenance is a significant challenge that DCAI has to consider to ensure data reliability in a dynamic environment. Our discussion involves data understanding, data quality assurance, and data acceleration.

\textbf{Data understanding} aims to develop algorithms and tools to help comprehensively understand data. One important measure is \emph{data visualization}~\cite{gathani2020debugging}, which represents data in a more intuitive form to facilitate communication. One notable example is the application of dimension reduction techniques, which can map complex and high-dimensional data onto a two-dimensional space, making it easier for humans to understand the data distribution. \emph{Data valuation} is another measure that helps understand what type of data is most valuable. It assigns credit to each data point regarding its contribution to the model performance. Understanding the data can help organizations make informed decisions about how to maintain the data effectively.

\textbf{Data quality assurance} is the key to model training. \emph{Quality assessment} aims to develop evaluation metrics to measure data quality and identify potential flaws and risks~\cite{pipino2002data}. \emph{Quality improvement} influences different stages of a data pipeline~\cite{loshin2010practitioner}. The improvement methods can be performed by domain experts (e.g., auditing and feedback), collective intelligence (e.g., majority voting), user-defined rules (e.g., data unit-test), or automated algorithms (e.g., correcting errors in labels).

\textbf{Data acceleration} aims to construct an efficient data infrastructure to facilitate the rapid acquisition of data. \emph{Resource allocation}~\cite{van2017automatic} is one way to achieve data acceleration, as it aims to effectively balance resources by managing database configurations to improve throughput and minimize latency. Another task is \emph{query acceleration}~\cite{pedrozo2018adaptive}, which aims to achieve rapid data retrieval by minimizing the number of disk accesses required during query processing and reducing workloads by identifying repeated queries. Data acceleration enables faster iterations of data developments.

Data maintenance is not an isolated component in DCAI but rather plays a fundamental and supporting role in the data ecosystem. With this perspective in mind, we predict that data maintenance strategies will become more dependent on training and evaluation data construction, providing them with continuous data support in an ever-changing environment.

\section{Open Research Challenges}
\label{sec:5}

To achieve DCAI, some open research challenges still deserve future exploration in our community.

\textbf{Inference Data \& Data Maintenance.} The majority of previous research has focused on training data development. The model-centric AI paradigm could partly drive this since many DCAI tasks were treated as pre-processing steps of model training. We argue that the role of data expands well beyond training, and inference data and data maintenance are equally important. These two comparatively underexplored directions are challenging since they can be open-ended and not as well defined as constructing training data. Rather than desperately optimizing performance metrics, they aim to provide a comprehensive understanding of the performance and continuous support of data.

\textbf{Cross-task Techniques.} Despite the progresses on various individual tasks, the investigation from the broader DCAI view is relatively lacking. In particular, different DCAI tasks could have an interaction effect. The optimal data augmentation choice may depend on the collected data; the evaluation set construction strategy needs to consider training data, and the training data could be adjusted based on the evaluation results; data maintenance strategies must be designed based on the training/evaluation data characteristics. It remains a challenge to systematically and simultaneously tackle data issues across multiple DCAI tasks. AutoML could be one of the promising directions to approach this goal with end-to-end data pipeline search~\cite{feurer2015efficient,heffetz2020deepline,feurer2020auto,lai2021tods,zha2021autovideo}.


\textbf{Data-model Co-design.} While DCAI advocates shifting our focus to data, it does not necessarily imply that the model has to stay unchanged. The optimal data strategies could differ when using different models and vice versa. With this in mind, our prediction is that future advancements will come from co-designing data pipelines and models and that the data-model co-evolvement will pave the way to more powerful AI systems. More studies are encouraged to investigate data-model relationships and co-design techniques.

\textbf{Data Bias.} Recently, AI systems in many high-stake applications have been reported to show discrimination against certain groups of people, which raises serious fairness concerns~\cite{du2019learning,wan2022processing,mehrabi2021survey}. The root cause often lies in the biased distributions of specific sensitive variables in data. From the DCAI perspective, some intriguing challenges arise: 1) How to mitigate the bias in the training data? 2) How to construct evaluation data to expose the unfairness issue? 3) How to continuously maintain data unbiasedness in a dynamic environment?

\textbf{Benchmarks.} In the model-centric paradigm, benchmarks propel us forward in advancing model designs. However, benchmarks are lacking for DCAI. The existing benchmarks often only focus on a specific DCAI task (e.g., feature selection~\cite{li2017feature}). It remains a challenge to construct a \emph{data benchmark} to understand the overall data quality and comprehensively evaluate various DCAI techniques~\cite{mazumder2022dataperf}. Efforts in DCAI benchmarking will significantly accelerate our research progress.

\vspace{-2pt}

\section{Conclusion}
\label{sec:6}
DCAI is an emerging research field that fundamentally shifts our focus from model to data. We provide a top-level discussion of its missions, aiming to help the community understand DCAI and push forward its progress. Many unsolved challenges remain to be addressed before we can fully achieve DCAI.

\vspace{-2pt}

\section*{Acknowledgements}
The work is, in part, supported by NSF (\#IIS-2224843). The views and conclusions in this paper are those of the authors and do not represent any funding agencies.

\vspace{-6pt}

\bibliographystyle{siamplain}
\bibliography{ref,fan_ref}

\end{document}